# An advantage actor-critic algorithm for robotic motion planning in dense and dynamic scenarios

Chengmin Zhou [1], Bingding Huang [2], Pasi Fränti [1,*]

[1] University of Eastern Finland, Joensuu, Finland

[2] Shenzhen Technology University, Shenzhen, China

[*] Correspondence: franti@cs.uef.fi

**Abstract:** Intelligent robots provide a new insight into efficiency improvement in industrial and service scenarios to replace human labor. However, these scenarios include dense and dynamic obstacles that make motion planning of robots challenging. Traditional algorithms like A* can plan collision-free trajectories in static environment, but their performance degrades and computational cost increases steeply in dense and dynamic scenarios. Optimal-value reinforcement learning algorithms (RL) can address these problems but suffer slow speed and instability in network convergence. Network of policy gradient RL converge fast in Atari games where action is discrete and finite, but few works have been done to address problems where continuous actions and large action space are required. In this paper, we modify existing advantage actor-critic algorithm and suit it to complex motion planning, therefore optimal speeds and directions of robot are generated. Experimental results demonstrate that our algorithm converges faster and stable than optimal-value RL. It achieves higher success rate in motion planning with lesser processing time for robot to reach its goal.

## I. Introduction

Intelligent robots are playing a crucial role in various scenarios of daily life. For example, luggage delivery in airport or food delivery in restaurant can use robot to replace human labor for improving the efficiency. These complex scenarios include many dynamic obstacles (e.g. pedestrians). The speed and moving directions of these obstacles are unpredictable, therefore it is challenging for robot to walk safely and efficiently towards the destination.

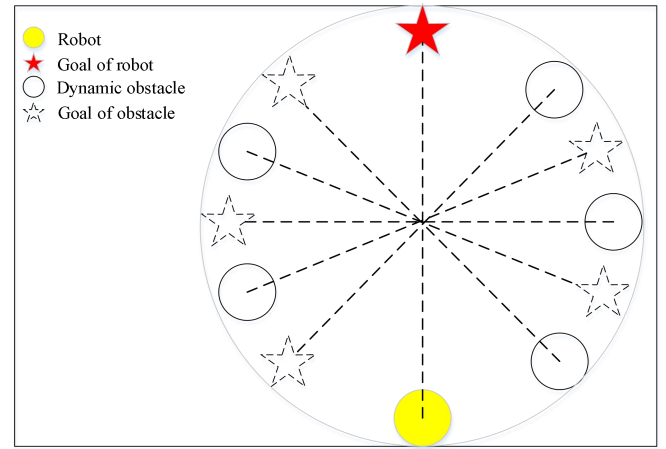

Figure 1. Circle-crossing scenario with dense and dynamic obstacles. Robot and all obstacles are in different velocities, and they are all generated in random positions within the circle.

Traditional graph search algorithms (e.g. A* [1], Dijkstra's algorithm [2]) and sampling-based algorithms (e.g. dynamic window approach [3], rapidly-exploring random tree [4]) can plan optimal and suboptimal collision-free trajectories to their destinations. These algorithms work well in static environment but it is challenging for robots to react properly in scenarios that include many dynamic obstacles, because these algorithms lack speed adjustment. Reciprocal velocity obstacles [5] work well in speed adjustment by computing a collision-free velocity to the goal. However, traditional algorithms mentioned above are all computationally expensive due to (1) large search space; (2) frequent updates of dynamic scenarios.

Recent reinforcement learning (RL) can address problems in traditional algorithms well by using a trained network (model) to directly plan the speed and direction simultaneously. Examples of RL algorithms mainly include optimal value RL (e.g. *Q* learning [6], deep *Q* network [7] and deep *V*-learning [8]) and policy gradient RL (e.g. asynchronous advantage actor-critic [9], deep deterministic policy gradient [10]). Their performance however varies. Network for outputting actions needs to be trained, but optimal value algorithm suffers slow speed and instability in network convergence because of over-estimation of action value in training [11], while policy gradient RL has better performance in convergence because of its actor-critic architecture [12]. However, actor-critic algorithms (e.g. asynchronous actor-critic) still suffer slow convergence speed in early-stage training because of its low-quality training set generated by their low-accuracy policy network in the beginning, although its speed is faster than optimal value algorithms.

Most policy gradient RL are widely used in games (e.g. Atari games) [13][22], in which outputted actions are finite and discrete, but only few algorithms can be found in scenarios where continuous actions and large action choices (action space) are required. This paper provides modifications to existing high-performance actor-critic algorithm to fit it to such scenarios that require continuous actions and large action space. This is achieved by (i) applying imitation learning [13][23][25] to enhance the speed and stability of network convergence; (ii) using selected experiences and experience replay memory [7] for offline training to improve speed of convergence; (iii) applying a novel network framework to advantage actor-critic algorithm.

## II. Problem formulation

Recent work [14] introduced a competent simulative environment (Figure 1) that includes dynamic robot and obstacles in a fix-size 2D indoor area. Robot and obstacles move towards their own goals simultaneously and avoid collisions with each other. Robot and obstacles obey the same or different policies for motion planning to avoid collisions. This simulative environment creates a circle-crossing scenario that adds predictable complexity to the environment. It is therefore a good platform to evaluate the performance of algorithms adopted by robot or obstacles.

Robot and obstacles plan motions towards their goals and avoid collisions by sequential decision making. Let $\boldsymbol{s}$ represents the state of robot. Let $\boldsymbol{a}$ and $\boldsymbol{v}$ represent action and velocity of robot, and $\boldsymbol{a} = \boldsymbol{v} = [v_x, v_y]$. Let $\boldsymbol{p} = [p_x, p_y]$ represents position of robot. let $\boldsymbol{s_t}$ represents state of robot at time step $t$. $\boldsymbol{s_t}$ is composed by observable and hidden parts $\boldsymbol{s_t} = [\boldsymbol{s_t^o}, \boldsymbol{s_t^h}]$, $\boldsymbol{s_t} \in \Re^9$. Observable part refers to factors of state that can be measured by others and is composed by position, velocity, and radius $\boldsymbol{s^o} = [p_x, p_y, v_x, v_y, r]$, $\boldsymbol{s^o} \in \Re^5$. Hidden part refers to factors of state that cannot be seen by others and is composed by planned goal position, preferred speed and heading angle $\boldsymbol{s^h} = [p_{gx}, p_{gy}, v_{pref}, \theta]$, $\boldsymbol{s^h} \in \Re^4$. State, position, and radius of obstacles are described by $\hat{\boldsymbol{s}}$, $\hat{\boldsymbol{p}}$ and $\hat{r}$.

To analyze this decision-making process, we first introduce one-obstacle one-robot case as in [8], and then extend it to one-robot multi-obstacle case. Robot plans its motion by obeying policy $\pi : (\boldsymbol{s_{0:t}}, \hat{\boldsymbol{s}}_{0:t}^o) \to \boldsymbol{a_t}$, while obstacles obey $\hat{\pi} : (\hat{\boldsymbol{s}}_{0:t}, \boldsymbol{s}_{0:t}^o) \to \boldsymbol{a_t}$. The objective of robot is to minimize the time to its goal $\mathbb{E}[t_g]$ (Eq. 1) under the policy $\pi$ without collisions to obstacles. Constraints of robot's motion planning can be formulated via Equations 2-5 that represent collision avoidance constraint, goal constraint, kinematics of robot and kinematics of obstacle respectively.

$$minimize. \; E[t_g | \boldsymbol{s_0}, \hat{\boldsymbol{s}}_0^o, \pi, \hat{\pi}] \quad (1)$$

$$s.t. \; \|\boldsymbol{p_t} - \hat{\boldsymbol{p}}_t\|_2 \geq r + \hat{r} \quad \forall t \quad (2)$$

$$\boldsymbol{p_{tg}} = \boldsymbol{p_g} \quad (3)$$

$$\boldsymbol{p_t} = \boldsymbol{p_{t-1}} + \Delta t \cdot \pi : (\boldsymbol{s_{0:t}}, \hat{\boldsymbol{s}}_{0:t}^o) \quad (4)$$

$$\hat{\boldsymbol{p}}_t = \hat{\boldsymbol{p}}_{t-1} + \Delta t \cdot \hat{\pi} : (\hat{\boldsymbol{s}}_{0:t}, \boldsymbol{s}_{0:t}^o) \quad (5)$$

Constraints of one-robot one-obstacle case can be easily extended into one-robot *N*-obstacle case where the objective (Eq. 1) is replaced by $minimize. \; E[t_g | \boldsymbol{s_0}, \{\hat{\boldsymbol{s}}_0^o ... \hat{\boldsymbol{s}}_N^o\}, \pi, \hat{\pi}]$ (assume that obstacles use the same policy $\hat{\pi}$). Collision avoidance constraint (Eq. 2) is replaced by

$$\begin{cases} \|\boldsymbol{p}_t - \widehat{\boldsymbol{p}}_{0:t}\|_2 \geq r + \hat{r} \\ \|\boldsymbol{p}_t - \widehat{\boldsymbol{p}}_{1:t}\|_2 \geq r + \hat{r} \\ \dots \\ \|\boldsymbol{p}_t - \widehat{\boldsymbol{p}}_{N-1:t}\|_2 \geq r + \hat{r} \end{cases} \forall t$$

assuming that obstacles are in the same radius $\hat{r}$. Kinematics of robot is replaced by $\boldsymbol{p}_t = \boldsymbol{p}_{t-1} + \Delta t \cdot \pi:(s_{0:t}, \{\hat{s}^o_{0:t} \dots \hat{s}^o_{N-1:t}\})$, and kinematics of obstacle is replaced by

$$\begin{cases} \widehat{\boldsymbol{p}}_{0:t} = \widehat{\boldsymbol{p}}_{0:t-1} + \Delta t \cdot \hat{\pi}:(\hat{s}_{0:t}, s^o_{0:t}) \\ \widehat{\boldsymbol{p}}_{1:t} = \widehat{\boldsymbol{p}}_{1:t-1} + \Delta t \cdot \hat{\pi}:(\hat{s}_{1:t}, s^o_{0:t}) \\ \dots \\ \widehat{\boldsymbol{p}}_{N-1:t} = \widehat{\boldsymbol{p}}_{N-1:t-1} + \Delta t \cdot \hat{\pi}:(\hat{s}_{N-1:t}, s^o_{0:t}) \end{cases}.$$

## III. Methods

A. Related works

**Traditional methods:** *graph search algorithm* (e.g. A* [1] and Dijkstra's algorithm [2]) fits constraints (Equations 2-5) by searching its working space using search tree or by exhaustive search (note that working space here refers to indoor space), but it can't fit Eq. 1 because velocity cannot be regulated in these algorithms, therefore expected time to goal is unpredictable. Additionally, the search process in graph search algorithms is computationally expensive by repeated search in the working space and computing an optimal trajectory. *Sampling-based algorithm* (e.g. dynamic window approach [3] and rapidly-exploring random tree [4]) samples the working space and computes a suboptimal trajectory. It fits constraints (Equations 2-5) but cannot fit Eq. 1 as graph search algorithm, and it is also computationally expensive especially in dynamic environment by frequently sampling working space and updating maps that are used for trajectory planning. *Reaction-based method* (e.g. optimal reciprocal collision avoidance (ORCA) [14]) fits the expected time to goal and constraints (Equations 1-5) well by computing one-step cost and selecting a suboptimal collision-free velocity to its goal. However, it is computationally expensive as it requires frequent updates to avoid dynamic obstacles.

**Optimal value RL:** *action-value-based RL algorithm* (e.g. deep Q network [7]) and *state-value-based RL algorithm* (e.g. deep V learning [8][21]) fit Equations 1-5 completely. Both are *optimal value RL algorithms* that are based on *Markov decision process* (MDP). Sequential decision problem defined by Equations 1-5 can be easily formulated as a MDP, which is described as a tuple $<S,A,P,R,\gamma>$. $S$ represents state and here refers to states of robot and obstacles. $A$ represents an action taken by robot. State $S$ transits into another state under a state-transition probability $P$ and a reward $R$ from the environment is obtained. $\gamma$ is a discount factor used for normalization of state value ($V$ value) or state-action value ($Q$ value). The objective of optimal-value-based RL is to find optimal $Q$ or $V$ values

$$Q^*(s,a) = \sum_{t=0}^{T} \gamma R(s, \pi^*(s,a)) \quad (6)$$
$$V^*(s) = \sum_{t=0}^{T} \gamma R(s, \pi^*(s)) \quad (7)$$

where $\gamma \in [0,1)$; $\pi^*(s,a)$ and $\pi^*(s)$ refer to optimal policies obtained by

$$\pi^*(s_t, a_t) = argmax_{a_t} R(s_t, a_t) +$$
$$\gamma \int_{s_{t+\Delta t}} P(s_t, s_{t+\Delta t}|a_t) Q^*(s_{t+\Delta t}, a_{t+\Delta t}) ds_{t+\Delta t} \quad (8)$$
$$\pi^*(s_t) = argmax_{a_t} R(s_t, a_t) +$$
$$\gamma \int_{s_{t+\Delta t}} P(s_t, s_{t+\Delta t}|a_t) V^*(s_{t+\Delta t}) ds_{t+\Delta t} \quad (9)$$

where $P(s_t, s_{t+\Delta t}|a_t)$ refers to the state-transition probability from time step $t$ to time step $t+\Delta t$. state-transition probability relates to kinematics of robot and obstacles, but it is unknown because motions of other obstacles are unpredictable. Optimal policy $\pi^*$, however, can still be represented using $Q^*$ or $V^*$ functions (Equations 8-9) by using neural network $\theta$ to approximate policy $\pi$, therefore the optimal action $a_t$ can be selected, for instance, by

$$a_t \leftarrow argmax_{a_t} R(s_t^{jn}, a_t) + \gamma^{\Delta t \cdot v_{pref}} V(s_{t+\Delta t}^{jn}; \theta) \quad (10)$$

as the line 7 in *Algorithm 1* [15]. State value $V$ is more suitable than action value $Q$ in action selection, because action space here is continuous and the set of permissible velocity vectors $\boldsymbol{v} = [v_x, v_y]$ depends on states of robot and obstacles (preferred speed) [8]. Note that action space here refers to all actions that can be possibly selected by robot or obstacles.

Optimal value RL algorithms, either state-value-based RL or action-value-based RL, are based on neural network to approximate $Q$ value or $V$ value. This process inevitably introduces error in value prediction [13], therefore leading to over-estimation of *Q* or *V* values. Over-estimation will lead to slow speed and instability of network convergence in training, therefore optimal-value-based RL are not the best choice to solve sequential decision problem defined by Equations 1-5.

---

**Algorithm 1: Deep *V*-learning**

1: Initialize value network *V* with demonstration *D*
2: Initialize target value network $\widehat{V} \leftarrow V$
3: Initialize experience replay memory $E \leftarrow D$
4: **For** episode = 1, M **do**
5:     Initialize random sequence $s_0^{jn}$
6:     **Repeat**
7:         $a_t \leftarrow argmax_{a_t \in A} R(s_t^{jn}, a_t) + \gamma^{\Delta t \cdot v_{pref}} V(s_{t+\Delta t}^{jn})$
8:         Store tuple $(s_t^{jn}, a_t, r_t, s_{t+\Delta t}^{jn})$ in *E*
9:         Sample random minibatch tuples from *D*
10:        Set target $y_i = r_i + \gamma^{\Delta t \cdot v_{pref}} \widehat{V}(s_{i+1}^{jn})$
11:        Update value network *V* by gradient descent
12:    **Until** terminal state $s_t$ or $t \geq t_{max}$
13:    Update target network $\widehat{V} \leftarrow V$
14: **End for**
15: **Return** *V*

---

**Policy gradient RL:** optimal value RL uses neural network to approximate *Q* or *V* value to indirectly select action via Eq. 10. *Policy gradient method* [26] uses neural network $\theta$ as policy $\pi_\theta: s \rightarrow a$ to directly select actions. Policy gradient method relays on trajectory reward $r(\tau)$ as Equation 11, instead of one-step reward in optimal value RL (note that $\tau$ refers to robot's interactive trajectory with the environment).

$$\pi^* = argmax_\pi E_{\tau \sim \pi(\tau)}[r(\tau)] \quad (11)$$

This direct and trajectory-reward-based way can reduce instability of network convergence in optimal-value-based RL. The objective of policy gradient method is defined as

$$J(\theta) = E_{\tau \sim \pi(\tau)}[r(\tau)] = \int_{\tau \sim \pi(\tau)} \pi_\theta(\tau) r(\tau) d\tau \quad (12).$$

Policy network $\theta$ can be updated with objective gradient $\nabla_\theta J(\theta)$ by

$$\theta \leftarrow \theta + \nabla_\theta J(\theta) \quad (13)$$

where $\nabla_\theta J(\theta) = \int_{\tau \sim \pi(\tau)} \nabla_\theta \pi_\theta(\tau) r(\tau) d\tau$. Trajectory can be divided into *T* step, therefore gradient value can be replaced by

$$\nabla_\theta J(\theta) = \sum_{t=0}^T \nabla_\theta log \pi_\theta(a_t|s_t) \cdot \sum_{t=0}^T r(s_t, a_t) \quad (14).$$

First part $\sum_{t=0}^T \nabla_\theta log \pi_\theta(a_t|s_t)$ is known by using neural network as approximator, therefore the gradient value relays on the second part $\sum_{t=0}^T r(s_t, a_t)$. However, using trajectory reward $\sum_{t=0}^T r(s_t, a_t)$ to update policy network inevitably slows down the speed of network convergence although stability is improved.

*Actor-critic algorithm* [12] is an optimized version of policy gradient method by using actor-critic architecture, in which one network (*policy network*) works as actor to select action and another network (*critic network*) works as critic to evaluate selected action to impact the update of policy network $\theta$. Policy network approximates policy value $\pi(a_t|s_t;\theta)$, while critic network approximates state value $V(s_t;\theta_v)$. Objectives of policy network and critic network are defined as

$$J(\theta) = \pi(a_t|s_t;\theta) \cdot [r_t + V(s_{t+1};\theta_v) - V(s_t;\theta_v)] (15)$$

$$J(\theta_v) = [r_t + V(s_{t+1};\theta_v) - V(s_t;\theta_v)]^2 \quad (16)$$

where $A = r_t + V(s_{t+1};\theta_v) - V(s_t;\theta_v)$ is the *temporal difference error* that measures the difference between *unbiased estimation* of $V(s_t)$ and predicted state value $V(s_t;\theta_v)$. Updates of policy and critic networks are obtained via Equations 17-18 as the update of policy gradient method (Eq. 13).

$$\theta \leftarrow \theta + \nabla_\theta log\pi(a_t|s_t;\theta) \cdot [r_t + V(s_{t+1};\theta_v) - V(s_t;\theta_v)] \quad (17)$$

$$\theta_v \leftarrow \theta_v + \nabla_{\theta_v}[r_t + V(s_{t+1};\theta_v) - V(s_t;\theta_v)]^2 \quad (18).$$

Actor-critic algorithm updates its networks ($\theta$ and $\theta_v$) using one-step reward or *N*-step rewards in interactive trajectory $\tau$, therefore speed of network convergence is improved. Denote that there is a trade-off in selecting *N*, because larger *N* will lead to slower convergence speed but stability is improved as policy gradient method.

Recent *asynchronous advantage actor-critic algorithm* (A3C) [9] and GPU-based A3C (GA3C) [16][20] move deeper in improving speed of network convergence by multi-thread method [9] to obtain more interactive trajectories and update its networks in each thread as the description in *Algorithm 2* [9]. However, existing work about *advantage actor-critic algorithm* (A2C) [9][17] shows that multi-thread method has limited contribution to speed of network convergence. Large portions of trajectories collected by either A3C or GA3C lack quality. For example, collected trajectories are with high quality if robot reaches the goal or collides with obstacles, while other trajectories collected under maximum time $t_{max} = t - t_{start}$ contribute less to speed of network convergence. Additionally, A3C and GA3C are online-learning algorithms. This means these algorithms cannot reuse collected high-quality trajectories in training, therefore also leading to low efficiency in network convergence. Moreover, these algorithms lack proper method in network initialization, therefore network converges slowly in early-stage training.

**Algorithm 2: A3C for each actor-learner thread**

//assume global shared parameter vectors $\theta$ and $\theta_v$
and global shared counter T = 0
//assume thread-specific parameter vectors $\theta'$ and $\theta'_v$
1: Initialize thread step counter $t \leftarrow 1$
2: **Repeat**
3:     Reset gradients: $d\theta \leftarrow 0$ and $d\theta_v \leftarrow 0$
4:     Synchronize thread-specific parameters:
        $\theta' \leftarrow \theta, \theta'_v = \theta_v$
5:     $t_{start} = t$
6:     Get state $s_t$
7:     **Repeat**
8:         Perform $a_t$ according to policy $\pi(a_t|s_t;\theta')$
9:         Receive reward $r_t$ and new state $s_{t+1}$
10:         $t \leftarrow t+1$
11:         $T \leftarrow T+1$
12:     **Until** terminal $s_t$ or $t - t_{start} = t_{max}$
13:     $R = \begin{cases} 0 & \text{for terminal } s_t \\ V(s_t, \theta'_v) & \text{for nonterminal } s_t \end{cases}$
14:     **For** $i \in \{t-1,...t_{start}\}$ **do**
15:         $R \leftarrow r_i + \gamma R$
16:         Accumulate gradients wrt $\theta'$:
        $d\theta \leftarrow d\theta + \nabla_{\theta'} \log\pi(a_i|s_i;\theta')(R - V(s_i;\theta'_v))$
17:         Accumulate gradients wrt $\theta'_v$:
        $d\theta_v \leftarrow d\theta_v + \partial(R - V(s_i;\theta'_v))^2 / \partial \theta'_v$
18:     **End for**
19:     Perform asynchronous update of $\theta$
20: **Until** $T > T_{max}$

B. Our optimized A2C for motion planning

Inspired by deep *V*-learning [8], A3C [9] and A2C [17], we modify A2C to solve the problem defined by Equations 1-5. We introduce an *A2C algorithm for motion planning* (A2CMP) shown in *Algorithm 3*.

The speed of network convergence in early-stage training is improved by imitation learning with high-quality demonstrations (interactive trajectories) *D* [13], which are generated by performing pretrained deep *V*-learning model [8]. Hence, policy network, value network (critic network) and their targeted networks are initialized (lines 1-3). Unlike online-learning algorithm (e.g. A3C), offline-learning method with experience replay memory $E$ is used in A2CMP to reuse collected high-quality demonstrations. Replay memory $E$ is initialized by running initialized policy network $\theta$ (line 4), and this is achieved by lines 6-15.

Experience is the tuple $<s_t^{jn}, a_t, V(s_t^{jn})>$ where $s_t^{jn}$ represents joint state composed by state of robot $s_t = [s_t^o, s_t^h]$ and observable state of obstacles $[\hat{s}_{0:t}^o, \hat{s}_{1:t}^o ... \hat{s}_{N-1:t}^o]$. Action $a_t$ is the vector $[v_x, v_y]$. State value $V(s_t^{jn})$ is composed by reward $r_t$ and predicted state value generated by preforming targeted critic network $\theta'_v$ (line 12)

$$V(s_t^{jn}) = r_t + \gamma^{\Delta t \cdot v_{pref}} V'(s_t^{jn}; \theta'_v) \quad (19)$$

where $\gamma \in [0,1)$ relates to time and preferred speed of robot. Reward $r_t$ is defined by

$$r_t = \begin{cases} -0.25 & \text{if } d < 0 \\ -0.1 - \frac{d}{2} & \text{else if } d < 0.2 \\ 1 & \text{else if } \boldsymbol{p} = \boldsymbol{p_g} \\ 0 & \text{else} \end{cases} \quad (20)$$

where $d$ represents the distance between robot and obstacles. Negative rewards are used to penalize robot if robot moves too close or collides with an obstacle [8]. Note that not all experiences are qualified for training, and qualified experiences are saved in replay memory $E$ when robot reaches the goal or collides with an obstacle (line 13). Interaction (episode) of robot with the environment terminates if robot reaches goal or collides with obstacles, or maximum interactive time is reached $t - t_0 = t_{max}$ (line 15).

Experience replay memory $E$ is updated in the same way as in initialization (lines 6-15). *N* random minibatch of tuple $<s_t^{jn}, a_t, V(s_t^{jn})>$ with length *L* are sampled from replay memory (line 16) for forthcoming training (lines 17-26). *Temporal difference error* (TD-error) is used as advantage *A* by

$$A = V(s_i^{jn}) - V'(s_i^{jn}; \theta'_v) \quad (21)$$

where $V(s_i^{jn})$ is the target state value obtained by Equation 19 and $V'(s_i^{jn}; \theta'_v)$ is the predicted state value by performing targeted critic network $\theta'_v$.

Objective of critic network is defined as

$$J(\theta'_v) = A^2 \quad (22).$$

Objective of policy network is defined as

$$J(\theta') = \pi(a_i|s_i^{jn};\theta') \cdot A + \beta H(\pi(s_i^{jn};\theta')) \quad (23)$$

where the first part $\pi(a_i|s_i^{jn};\theta') \cdot A$ is the same as the objective of policy network of A3C [9], while the second part $\beta H(\pi(s_i^{jn};\theta'))$ is the *policy entropy* [18] that can encourage exploration of better actions in early-state training for speeding up network convergence. Here $\beta \in (0,1]$ is a discount factor to balance the exploration. According to objectives of critic and policy networks, losses of critic and policy networks are defined by

$$\mathcal{L}_v = \partial A^2 / \partial \theta' \quad (24)$$

$$\mathcal{L}_\pi = \nabla_{\theta'} log\pi(a_t|s_i^{jn};\theta') \cdot A + \beta \nabla_{\theta'} H(\pi(s_i^{jn};\theta')) \quad (25).$$

Targeted networks of policy and critic networks are updated by minimizing total loss $\mathcal{L}_{total}$

$$\mathcal{L}_{total} = \mathcal{L}_\pi + \lambda \mathcal{L}_v \quad (26)$$

where $\lambda \in (0,1]$ and it is a discount factor that is used to balance the update speed of targeted critic network. Critic network and policy network are updated after *K* episodes of training (line 27), and policy network is saved after *M* episodes of training (line 29).

**Algorithm 3: A2C for motion planning (A2CMP)**
1: Collect demonstration D by pretrained deep *V*-learning model
2: Initialize policy network $\theta$ and value network $\theta_v$ by imitation learning with demonstration D
3: initialize target policy network and target value network:
$\theta' \leftarrow \theta, \theta'_v \leftarrow \theta_v$
4: Initialize experience replay memory E
5: **For** episode = 1, M **do**
6:     Initialize random sequence $s_0^{jn}$
7:     **Repeat**
8:         Perform $a_t$ according to policy network $\theta$
9:         Receive reward $r_t$
10:        Save experience tuple $<s_t^{jn},a_t,r_t>$
11:        Get predicted state value $V'(s_t^{jn};\theta'_v)$
12:        Set target state value:
$V(s_t^{jn}) = r_t + \gamma^{\Delta t \cdot v_{pref}} V'(s_t^{jn};\theta'_v)$
13:        Update $E < s_t^{jn},a_t,V(s_t^{jn}) >$ if terminal state $s_t^{jn}$
14:        $t \leftarrow t + 1$
15:        **Until** terminal state $s_{terminal}^{jn} = \begin{cases} reachgoal \\ collision \end{cases}$ or $t - t_0 = t_{max}$
16:     Sample *N* random minibatch $< s_i^{jn},a_i,V(s_i^{jn}) >$ with size *L* from *E*
17:     **For** $i \in \{1...L\}$ **do**
18:        Get predicted state value $V'(s_i^{jn};\theta'_v)$
19:        Assume advantage value:
$A = V(s_i^{jn}) - V'(s_i^{jn};\theta'_v)$
20:        Set critic objective:
$J(\theta'_v) = A^2$
21:        Set policy objective:
$J(\theta') = \pi(a_i|s_i^{jn};\theta') \cdot A + \beta H(\pi(s_i^{jn};\theta'))$
22:        Get critic loss: $\mathcal{L}_v = \partial A^2/\partial \theta'$
23:        Get policy loss:
$\mathcal{L}_\pi = \nabla_{\theta'} log\pi(a_t|s_i^{jn};\theta') \cdot A + \beta \nabla_{\theta'} H(\pi(s_i^{jn};\theta'))$
24:        Get total loss: $\mathcal{L}_{total} = \mathcal{L}_\pi + \lambda \mathcal{L}_v$
25:        update networks:
$\theta' \leftarrow \theta'$ and $\theta'_v \leftarrow \theta'_v$ by minimizing total loss $\mathcal{L}_{total}$
26:     **End for**
27:     update networks:
$\theta \leftarrow \theta'$ and $\theta_v \leftarrow \theta'_v$ every K episode
28: **End for**
29: Return policy network $\theta$

## IV. Experiment

A. Design of neural network

Four linear layers are used in framework of A2CMP: *linear 1* as an input layer, *linear 2* as a hidden layer, *linear critic* as an output layer of critic network and *linear actor* as an output layer of policy network. Two *ReLU* layers are used as activation after linear 1 and linear 2 layers. One *softmax* layer is used for activation and normalization, therefore features are mapped to probability distribution. Framework of neural network is shown in Figure 2.

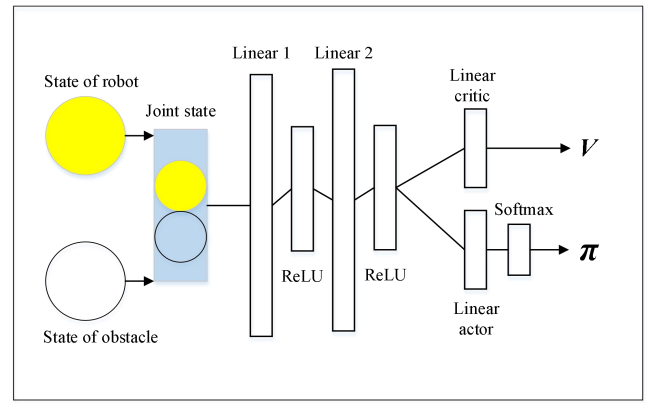

Figure 2. Network framework of A2CMP. It shows one-robot one-obstacle case where the joint state $[s,\hat{s}^o]$ is composed by full state of robot and observable state of obstacle. Experiment includes five obstacles therefore joint state is

$[s, \{\hat{s}_0^o, \hat{s}_1^o, \hat{s}_2^o, \hat{s}_3^o, \hat{s}_4^o\}]$.

Input is the joint state that is composed by full state $s = [p_x, p_y, v_x, v_y, r, p_{gx}, p_{gy}, v_{pref}, \theta]$, $s \in \Re^9$ and observable state of obstacles $\hat{s}^o = [p_x, p_y, v_x, v_y, r] \times N$, $\hat{s}^o \in \Re^5$ where $N$ is the number of obstacles. Here we choose $N=5$ for our experiments. Hence, the input size is 9+5×N. Number of directions is set to 16 ($directions = [i/16 \cdot 2\pi], i \in \{1,2,.16\}$), and number of sampled speeds is set to 5 ($speeds = [i/5 \cdot v_{pref}], i \in \{1,2,.5\}$). Hence, action space is of size 16×5+1 where 1 represents zero speed, therefore output size of linear actor is 16×5+1. Size of each linear layer is summarized in Table 1: (9+5×N, 128), (128, 256), (256, 1) and (256, 16×5+1) for linear 1, linear 2, linear critic and linear actor respectively. Format of outputted action after linear actor is a 2-dimension vector $\boldsymbol{a} = [v_x, v_y]$ that cannot be directly used for normalization in softmax layer. According to one-hot encoding, we use the label of each outputted action in action space to represent outputted action for normalization. Hence, outputs of softmax are labels of action that will be parsed to actions by mapping labels to actions.

Table 1. Size of network layers.

| Name | Size (input size, output size) |
|---|---|
| Linear 1 | (9+5×N, 128) |
| Linear 2 | (128, 256) |
| Linear critic | (256, 1) |
| Linear actor | (256, 16×5+1) |

B. Training

Experiences for training are tuples $<s^{jn}, \boldsymbol{a}, V(s^{jn})>$ from qualified trajectories collected from simulative environment shown in Figures 1 and 3. Simulative environment is the circle-crossing simulator from RVO2 [14]. Width of the circle is set to 8m and position of circle's center is set to (0, 0). Radius $r$ of robot and obstacles are set to 0.3-0.5m, and 5 obstacles are in the same radius. Preferred speed $v_{pref}$ is set to 1.0 m/s. Position of goal $[p_{gx}, p_{gy}]$ is randomly set within the circle, and starting position $[p_x, p_y]$ of robot is on the opposite position within circle. Goal positions of obstacles $[\hat{p}_{gx}, \hat{p}_{gx}]$ are also randomly set, and starting positions of obstacles $[\hat{p}_x, \hat{p}_x]$ are also on the opposite position within circle.

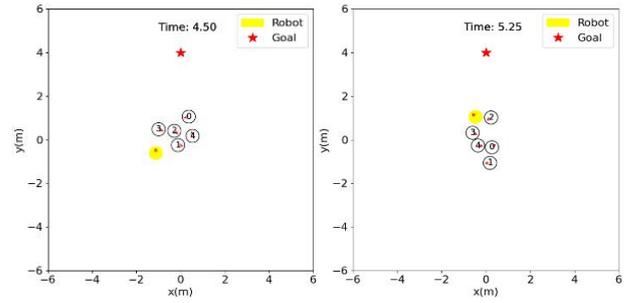

Figure 3. Circle-crossing simulator in the experiment.

Capacity of experience replay memory $E$ is set to 100000. Number of demonstrations $D$ collected from deep $V$-learning [8] for imitation learning is set to 3000 episodes. Learning rate and epochs of imitation learning are set to 0.01 and 50 respectively. Learning rate, batch size and training episodes $M$ of A2CMP are set to 0.001, 100 and 10000 respectively. Critic network and policy network are updated every 50 episodes. Number of sampled minibatches for training is set to 1. Entropy coefficient $\beta$ and critic loss coefficient $\lambda$ are set to 0.01 and 0.5 respectively. Values of all parameters are summarized in Table 2.

Table 2. Values of parameters used for training.

| Parameters | value |
|---|---|
| Number of demonstrations $D$ | 3000 |
| Imitation learning rate | 0.01 |
| Imitation epochs | 50 |
| A2CMP learning rate | 0.001 |
| Batch size $L$ | 100 |
| Training episodes $M$ | 10000 |
| Policy update interval $K$ | 50 |
| Memory capacity | 100000 |
| Sampled minibatch $N$ | 1 |
| Entropy coefficient $\beta$ | 0.01 |
| Critic loss coefficient $\lambda$ | 0.5 |

C. Evaluation

Deep $V$-learning and *A2CMP without imitation learning* are used as control groups to compare speed of convergence and stability of training with *A2CMP with imitation learning*. Networks of these algorithms are all trained by 10000 episodes of experience. Network of deep $V$-learning converges after about 8000 episodes. It converges slowly with small average rewards in the early-stage training. A2CMP network without imitation learning converges after about 6000 episodes, and it receives higher average rewards than deep *V*-learning. A2CMP network with imitation learning converges near 4000 episodes and average rewards received by A2CMP are higher than deep *V*-learning and A2CMP without imitation learning. A2CMP network converges more stable, while networks of deep *V*-learning and A2CMP without imitation learning lack stability according to their fluctuated training curves shown in Figure 4. Hence, we can conclude that A2CMP is more stable than deep *V*-learning in training and it uses less training experiences than deep *V*-learning in network convergence. We also see that imitation learning for network initialization contributes a lot to stability and speed of convergence in training.

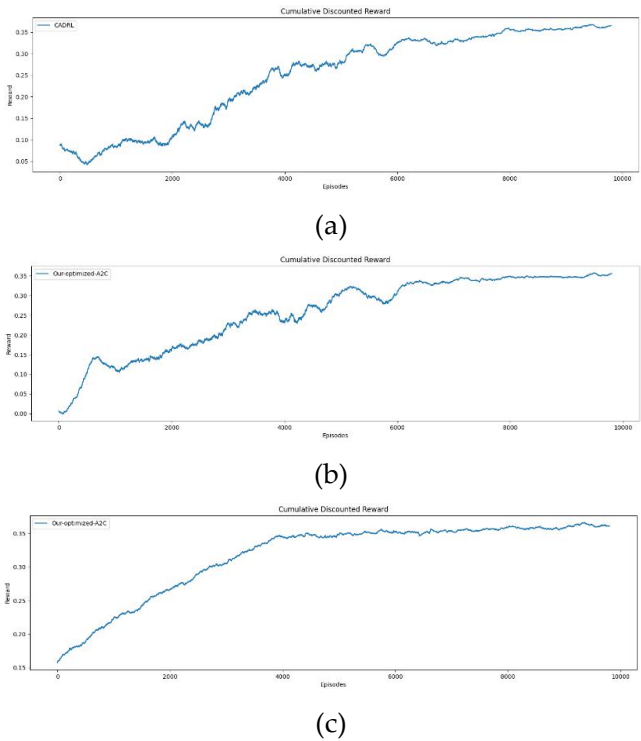

(a)

(b)

(c)

Figure 4. Training curves. (a) represents average rewards received by deep *V*-learning. (b) represents average rewards received by A2CMP without imitation learning. (c) represents average rewards received by A2CMP.

Performance of A2CMP is evaluated in circle-crossing simulative environment where starting and goal positions of robot are fixed at (-4, 0) and (4, 0) respectively. Positions of obstacles are randomly set within the circle, while their goal positions are always on the opposite position within circle. ORCA [14] and deep *V*-learning are used as control groups for comparisons. Trained networks of these algorithms are all evaluated by 100 testing episodes. Each episode terminates once robot reaches goal or experimental time reaches maximum testing time $t_{max} = 25s$. Trajectories and positions of robot and obstacles at each time step are shown in Figure 5. Success rate of ORCA after 100 testing episodes is 0.99 which is higher than that of deep *V*-learning (0.95) and A2CMP (0.96). Average time to reach goal by A2CMP (10.2s) is slightly better than that of deep *V*-learning (10.4s), and it is about 17% shorter than the time spent by ORCA (12.5s).

Rates of collision and goal missing by deep *V*-learning and A2CMP are higher than that by ORCA. This may relate to the defective expression of joint state $s^{jn}$ that simply combines robot's state and obstacles' state together. We cannot completely represent relationships of robot and dynamic obstacles in this way. It could be fixed by using a new state expression (e.g. attention weight [15] and relation graph [19]), but it is beyond the scope of this paper. Detailed comparisons of these algorithms are summarized in Table 3.

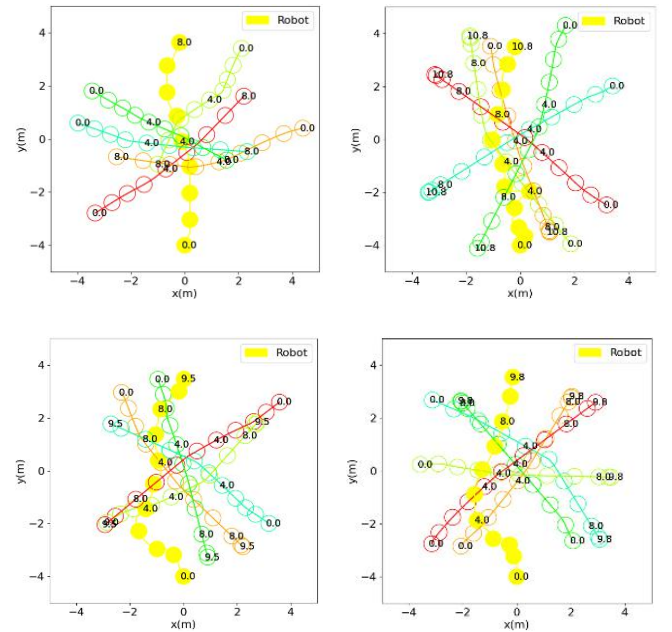

Figure 5. Trajectory examples in evaluation. Figure lists four cases within 100 evaluations. Yellow circle represents positions of the robot, and other colors denote that of obstacles at each time step. Time to reach goals varies because the complexity of each case differs. Average time to reach goal is therefore used to evaluate the performance of algorithms.

Table 3. Performance comparisons in evaluations.

| Algorithm | Success rate | Collision rate | Goal missing* | Average time (s) |
|-----------|--------------|----------------|---------------|------------------|
| ORCA      | 0.99         | 0.00           | 0.01          | 12.5             |
| DVL       | 0.95         | 0.02           | 0.03          | 10.4             |
| **A2CMP** | **0.96**     | **0.02**       | **0.02**      | **10.2**         |

*Goal missing represents that robot fails to reach goal when experimental time reaches the maximum time.

## V. Conclusion and future works

Our A2CMP algorithm is the modified version of advantage actor-critic algorithm, in which actor-critic architecture contributes to speed of network convergence compared to deep *V*-learning algorithm. A2CMP network learns from experience offline by using experience replay memory to store high-quality experiences, therefore speeding up the convergence of network. Imitation learning is also used in network initialization, and it contributes to convergence stability and convergence speed simultaneously in training. Results of training and evaluation demonstrate that our A2CMP algorithm converges faster and is more stable than deep *V*-learning. A2CMP costs less time to reach goal compared to ORCA and deep V learning, but a few collisions and goals missing still exist.

Future works will concentrate on better ways in state expression for a better description of robot and dynamic obstacles. For example, future motions of dynamic obstacles should be predicted and used as the feature in new state expression by estimating from their previous trajectories. In this way, algorithms can learn more from the state and cases like collision or goal missing can be better avoided.

## Acknowledgment

Thanks for open-source implementations of ROV2 and deep *V*-learning provided by [14][24] and [15].